\def\year{2021}\relax
\documentclass[letterpaper]{article} 
\usepackage{aaai21}  
\usepackage{times}  
\usepackage{helvet} 
\usepackage{courier}  
\usepackage[hyphens]{url}  
\usepackage{graphicx} 
\usepackage{natbib}  
\usepackage{caption} 
\usepackage{booktabs}
\usepackage{amsmath}
\usepackage{amsthm}
\usepackage{amsfonts}
\usepackage{multirow}
\usepackage{comment}
\usepackage{enumitem}
\usepackage[switch]{lineno}
\usepackage{caption}
\usepackage{subcaption}
\usepackage[symbol]{footmisc}

\newenvironment{proofsketch}{%
  \proof}{\endproof}

\newcommand{\meqref}[1]{Eq. \eqref{#1}}
\newcommand{\mtabref}[1]{Table \ref{#1}}
\newcommand{\mfigref}[1]{Figure \ref{#1}}

\newcommand{\mdata}{Taobao Dataset}
\newcommand{\mus}{Taobao}

\newcommand{\modelname}{Elapsed-Time Sampling Delayed Feedback Model}
\newcommand{\modelnameabb}{ES-DFM}

\title{Capturing Delayed Feedback in Conversion Rate Prediction \\
via Elapsed-Time Sampling}
\author {
        Jia-Qi Yang\textsuperscript{\rm 1}\footnote{Jia-Qi Yang performed this work as an intern at Alibaba.},
        Xiang Li\textsuperscript{\rm 2}\footnotemark[2],
        Shuguang Han\textsuperscript{\rm 2},
        Tao Zhuang\textsuperscript{\rm 2}\\
        De-Chuan Zhan\textsuperscript{\rm 1}\footnote{De-Chuan Zhan and Xiang Li  are the corresponding authors. This work was supported by NSFC (61773198, 6163000043, 61921006, 61751306).},
        Xiaoyi Zeng\textsuperscript{\rm 2},
        Bin Tong\textsuperscript{\rm 2}\\
}
\affiliations {
    \textsuperscript{\rm 1} State Key Laboratory for Novel Software Technology, Nanjing University, Nanjing, China \\
    \textsuperscript{\rm 2} Alibaba Group, Hangzhou, China\\
    \textsuperscript{\rm 1}\{yangjq,zhandc\}@lamda.nju.edu.cn\\
    \textsuperscript{\rm 2}\{leo.lx,shuguang.sh,zhuangtao.zt,yuanhan,tongbin.tb\}@alibaba-inc.com
}

\begin{document}


\maketitle

\begin{abstract}

Conversion rate (CVR) prediction is one of the most critical tasks for digital display advertising. Commercial systems often require to update models in an online learning manner to catch up with the evolving data distribution. However, conversions usually do not happen immediately after user clicks. This may result in inaccurate labeling, which is called delayed feedback problem. In previous studies, delayed feedback problem is handled either by waiting positive label for a long period of time, or by consuming the negative sample on its arrival and then insert a positive duplicate when conversion happens later. Indeed, there is a trade-off between waiting for more accurate labels and utilizing fresh data, which is not considered in existing works. To strike a balance in this trade-off, we propose \modelname{} (\modelnameabb{}), which models the relationship between the \textit{observed conversion} distribution and the \textit{true conversion} distribution. Then we optimize the expectation of \textit{true conversion} distribution via importance sampling under the elapsed-time sampling distribution. We further estimate the importance weight for each instance, which is used as the weight of loss function in CVR prediction. To demonstrate the effectiveness of \modelnameabb{}, we conduct extensive experiments on a public data and a private industrial dataset. Experimental results confirm that our method consistently outperforms the previous state-of-the-art results.

\end{abstract}

\section{Introduction}

Digital display advertising has become the main business model for many online services, in which advertisers pay for placing ads on those platforms. Among the available payment options, paying per conversion (CPA) is usually the dominated mechanism as conversions can directly bring profits. In the CPA model, advertisers pay only if users performed certain predefined conversion actions with the advertisement. To effectively display ads, machine learning models have been widely adopted to forecast the conversion rate (CVR), which is widely investigated in both academia and industry~\cite{lee2012estimating,chapelle2014simple,ma2018entire}. 

In order to capture the dynamic change of user needs, commercial systems often update learned models with up-to-date data within a short time, i.e., in an online training manner~\cite{jugovac2018streamingrec,guo2019streaming,FNW}. This further complicates the CVR prediction since conversions usually do not happen immediately after user clicks. The \textit{Delayed Feedback} issue introduces a dilemma for streaming CVR prediction --- on the one hand, we need to wait for a sufficient long time so that the observation information can approximately reflect the true conversion; on the other hand, we also tend to update the learned models without much delay for model-freshness. 

~\citet{DFM} was among the early studies to address the delayed feedback problem. The proposed Delayed Feedback Model (DFM) optimizes CVR as a joint probability over the predicted CVR and the delayed time distribution. This joint probability is estimated in the observed time interval, which may be biased from the true conversion distribution. The biased CVR is probably more inaccurate due to the delayed feedback problem in online learning settings. 

To achieve unbiased CVR estimation in delayed feedback problem, recent studies have explored the way of optimizing the expectation of true conversion distribution via importance sampling~\cite{importance_weight}.~\citet{FNW} proposed the Fake Negative Weighted (FNW) approach, in which each arriving instance is firstly labeled as negative, and then corrected upon its conversion at a later time. Each fake negative instance may have a side-effect for the learned model until it is amended. This side-effect is amplified if the data distribution frequently changes. For example, in the beginning of a promotion event, user clicks may increase dramatically while most conversions come after a certain time. Such overwhelming fake negatives may harm the predictive model. Instead of blindly labeling each incoming example as a negative instance,~\citet{FSIW} proposed a Feedback Shift Importance Weighting (FSIW) algorithm, in which the model waits for the real conversion in a certain time interval. However, FSIW does not allow the data correction even a conversion event took place afterward. We argue that positive examples are important for delayed feedback prediction as the positive examples are always more scarce than the negative examples. Moreover, FSIW may lack model-freshness due to a long-time wait. Therefore, either updating the model in nearly real time~\cite{FNW}, or waiting a sufficiently long time for conversion ~\cite{FSIW} may not be able to address the delayed feedback problem in the streaming CVR prediction. 

For unbiased CVR estimation in the online settings, we propose to wait for a time interval which is modeled as a distribution. The readily available conversion information allows the model to trade-off the label correctness and online model-freshness, which are achieved in FSIW and FNW, respectively. Due to the introduction of the observed time distribution, delayed positive samples can be better handled than FNW via important sampling techniques. Especially in scenarios of promotion events, FNW may fail to have unbiased estimation due to that the distribution of positive samples in observed time may be dramatically different from the routines. On the other hand, FSIW is able to guarantee the label correctness but lacks of model-freshness. Furthermore, the model is not able to correct instance label even the delayed positive instance comes later. The introduction of time distribution in our proposal helps the model correct the label of an instance by degrading the weight of negative instance and upgrading the weight of positive instance. 

In this work, we propose \modelname{} (\modelnameabb{}), which models the relationship between the \textit{observed conversion} distribution and the \textit{true conversion} distribution. We optimize the expectation of \textit{true conversion} distribution via importance sampling under the elapsed-time sampling distribution. We further estimate the importance weight for each instance, which is used as the weight of loss function in CVR prediction. To demonstrate the effectiveness of \modelnameabb{}, we conduct extensive experiments on two widely-used datasets --- a public ads conversion logs provided by Criteo, and a private data set provided by \mus. Experimental results confirm that our method consistently outperforms the previously state-of-the-art results in most of the cases. Our main contribution can be summarized as following:

\begin{itemize}
\item{To the best of our knowledge, we are the first to study the trade-off between waiting more accurate labels and exploiting fresher training data in the context of streaming CVR prediction.}

\item{By explicitly modeling the elapsed time as a probability distribution, we achieve the unbiased estimation of true conversion distribution. Particularly, our model is shown to be robust even if the data distribution is different from the routines.}

\item{We provide a set of rigorous experimental setups for streaming training and evaluation, which better aligns with industrial systems, and can be easily applied to real-world applications.}

\end{itemize}

\section{Related Work}

\subsection{Delayed Feedback Models} 
The mostly cited work that were addressing the delayed feedback problem came from ~\citet{DFM}, in which the authors stated that such a problem is related to the survival time analysis~\cite{survival_time}. The Delayed Feedback Model (DFM) assumed an exponential delay for conversion time distribution, and, based on that, proposed two models: one focusing on CVR prediction and the other on conversion delay prediction. 
Built on top of the DFM model, ~\citet{npdfm} further proposed a non-parametric delayed feedback model (NoDeF), in which the delay time was modeled without any parametric assumptions. One significant drawback of the above methods was that both of them only attempted to optimize the observed conversion information rather than the actual delayed conversion. 

\subsection{Importance Sampling}
Using samples from one distribution to estimate the expectation with respect to another distribution can be achieved by importance sampling method. ~\citet{FNW} proposed fake negative weighted method (FNW) to optimize the ground truth CVR prediction objective based on importance sampling. Under the assumption that all samples are initially labeled as negative, the delayed feedback problem can be resolved by FNW in expectation. However, in the streaming setting, every fake negative will affect the model negatively until it's corresponding positive duplicate arrives. This negative effect can be amplified drastically under distribution change. ~\citet{FSIW} proposed a feedback shift importance weighting method (FSIW), where the importance weight is estimated with the aid of waiting time information. However, FSIW does not allow duplicated samples, thus cannot correct the mislabeled samples using the subsequent positive labels by inserting duplicates.

\subsection{Delayed Bandits}
The delayed feedback in the bandit algorithm has been researched~\cite{joulani2013online,mandel2015queue,cesa2019delay}. The aforementioned approaches often provide efficient and provably optimal algorithms for delayed feedback scenarios. However, such methods naturally wait for having received enough
feedback before actually learning something, which may be quite unsuitable in a non-stationary environment.~\citet{vernadenon} defined a new stochastic bandit model and addressed the real-world modelling issues when dealing with non-stationary environments and delayed feedback. However, the objective in the bandit problem is to sequentially make decisions in order to minimize the cumulative regret, our goal is to predict the CVR in order to derive a bid price in ad auction.

\section{Background}

\begin{figure}
\centering
    \includegraphics[width=\linewidth]{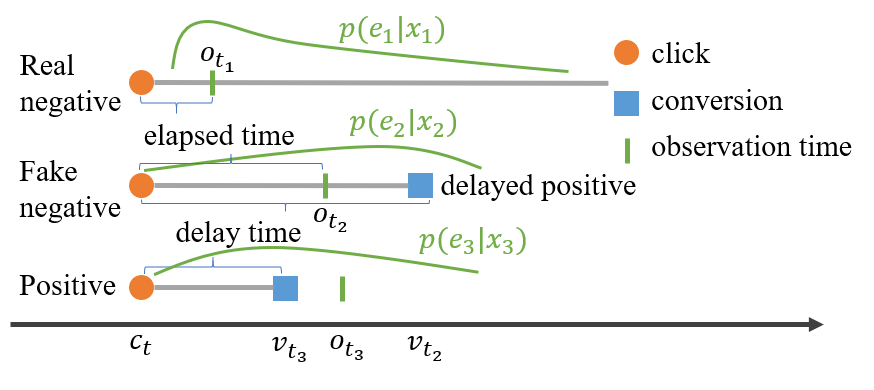}
    \caption{An illustration of different kind of time information for the delayed feedback tasks.}
    \label{fig:delayed-feedback-time-illustration}
\end{figure}

In this work, we focus on the CVR prediction task, which takes the user features $x_u$ and the item features $x_i$ as inputs, all the features are denoted by $x$, and aims to learn the probability that the user converts on the item. $y\in \{0, 1\}$ indicates the conversion label, where $y=1$ means the conversion, otherwise $y=0$. Ideally, the CVR model is trained on top of the training data $(x, y)$ drawn from the data distribution of ground-truth $p(x, y)$, thereby optimizing the ideal loss shown as follows:

\vspace{-3mm}
\begin{equation}
    \mathcal{L}_{ideal} = \mathbb{E}_{(x, y)\sim p(x, y)} \ell(y, f_\theta(x))
\end{equation}
where $f$ is the CVR model function, and $\theta$ is the parameter. $\ell$ is the classification loss, and the widely-used cross entropy is adopted. However, due to the delayed feedback problem, the observed distribution of the training data $q(x, y)$ often deviates from the distribution of the ground-truth $p(x, y)$. Therefore, the ideal loss $\mathcal{L}_{ideal}$ is unavailable.

To formulate such a delayed feedback setting more precisely, we introduce three time points and corresponding time intervals in Figure \ref{fig:delayed-feedback-time-illustration}. These three time points are the click time $c_t$ when a user clicks an item, the conversion time $v_t$ when a conversion action happens, and the observation time $o_t$ when we extract the training samples. Then the time interval between $c_t$ and $o_t$ is denoted as the \textit{elapsed time} $e$, and the time interval between $c_t$ and the $v_t$ is denoted as the \textit{delayed feedback time} $h$. Therefore, the samples are labeled as $y=1$ (positive) in the training data, when $e > h$, otherwise some positive samples are mislabeled as $y=0$ (fake negative) when $e < h$.

\section{Proposed Method} \label{sec::method}

In order to realize flexible control on the waiting time, we assume the elapsed time is drawn from an elapsed time distribution $p(e|x)$. Then we developed a probabilistic model that combines the elapsed time distribution $p(e|x)$, the delay time distribution $p(h|x, y=1)$ and the conversion rate $p(y=1|x)$ into a unified framework. To achieve an unbiased estimation of the actual CVR prediction objective, we propose an importance weighting method corresponding to our elapsed sampling method. Then we provide a practical estimation of the importance weights, and give an analysis of the bias introduced by this estimation, which can guide us on designing an appropriate elapsed time distribution $p(e|x)$.

\subsection{\modelname{}}
To strike the balance between obtaining accurate feedback information and keeping model freshness, a reasonable waiting time (elapsed time) should be integrated into the modeling process. Moreover, the elapsed time $e$ should be a distribution depend on $x$, i.e, $p(e|x)$. For example, users need more time to consider when buying high-priced products, thus a long waiting time is required. When a click $x_i$ arrives, an elapsed time $e_i$ is drawn from $p(e|x_i)$. Then we wait the sample $x_i$ for the time interval of $e_i$ before assigning a label, and subsequently train on the data. By introducing the time distribution, we propose our \modelname{} (\modelnameabb{}), which modeling the relationship between the \textit{observed conversion} distribution  $q(y|x)$ and the \textit{true conversion} distribution $p(y|x)$, according to:

\begin{equation}\label{gdfm0}
\begin{aligned}
    q(y=0|x) &= p(y=0|x) \\
    &+ p(y=1|x)p(h > e|x, y=1)
\end{aligned}
\end{equation}

\begin{equation}\label{gdfm1}
    q(y=1|x) = p(y=1|x)p(h \le e|x, y=1)
\end{equation}
where 
\begin{equation}
\begin{aligned}
    &p(h > e|x, y=1)\\
    &=\int_0^\infty \left[ p(e|x) \int_e^\infty p(h|x, y=1)dh\right]de
\end{aligned}
\end{equation}

\begin{equation}
\begin{aligned}
&p(h \le e|x, y=1)\\
&=\int_0^\infty \left[ p(e|x) \int_0^e p(h|x, y=1)dh\right]de
\end{aligned}
\end{equation}

At the time of model training, some conversions that will occur eventually have not yet been observed, and previous methods like DFM and FSIW have ignored these conversions. We argue this is important for a delayed feedback task as the positive examples are way more scarce than the negative examples, and the positives may define the direction for model optimization. Therefore, in this work, as soon as the user engages with the ad, the data will be sent (duplicated if there is already a fake negative) to the model with a positive label. Then, $q(y|x)$ should be re-normalized as following:
\vspace{-0mm}
\begin{equation}
    q(y=0) = \frac{p(y=0) + p(y=1)p(h > e|y=1)}{1+p(y=1)p(h > e|y=1)}\label{dp_gdfm1}
\end{equation}
\vspace{-0mm}
\begin{equation}
    q(y=1) =\frac{p(y=1)}{1+p(y=1)p(h > e|y=1)}\label{dp_gdfm2}
\end{equation}
where the condition on $x$ is omitted for conciseness, i.e. $q(y=0)=q(y=0|x)$, $p(y=0)=p(y=0|x)$, etc. Since we have inserted delayed positives, the total number of samples will increase by $p(y=1)p(h > e|y=1)$, so we should normalize by dividing $1+p(y=1)p(h > e|y=1)$. The number of negatives will not change, so dividing \meqref{gdfm0} by this normalizing factor yields \meqref{dp_gdfm1}. The number of positives will increase by $p(y=1)p(h > e|y=1)$, so the numerator of $q(y=1)$ is $p(y=1)p(h <= e|y=1)+p(y=1)p(h > e|y=1)$. Using the fact that $p(h <= e|y=1)+p(h > e|y=1)=1$ yields \meqref{dp_gdfm2}.

\subsection{Importance Weight of \modelnameabb}
To obtain unbiased CVR estimation in delayed feedback problem, we optimize the expectation of $p(y|x)$ via importance sampling~\cite{importance_weight}. First, we provide the theoretical background of importance sampling, as follows:
\begin{align}
    \mathcal{L}_{ideal} &= \mathbb{E}_{(x, y)\sim p(x, y)} \ell(y, f_\theta(x))\label{is_1}\\
    &= \int p(x)dx \int p(y|x) \ell(y, f_\theta(x))dy\\
    &= \int p(x)dx \int q(y|x)\frac{p(y|x)}{q(y|x)} \ell(y, f_\theta(x))dy\label{is_3}\\
    &\approx \mathbb{E}_{(x, y) \sim q(x, y)} \frac{p(y|x)}{q(y|x)}\ell(y, f_\theta(x))\label{is_4} \\
    &= \mathcal{L}_{iw}\label{is_last}
\end{align}
where $f$ is the CVR model function, and $\theta$ is the parameter. $\ell$ is the classification loss, and the widely-used cross entropy is adopted. Notice that we assume $p(x)\approx q(x)$ to obtain \eqref{is_4} from \eqref{is_3}, which is reasonable since the proportion of delayed positive is small, and this approximation is also used by \citet{FNW}. According to \eqref{is_4}, we can optimize the ideal objective with a appropriate weight $w(x,y)=\frac{p(y|x)}{q(y|x)}$.  Second, we further provide the importance weight under the proposed elapsed-sampling distribution. From Equation \eqref{dp_gdfm1} and \eqref{dp_gdfm2}, we can obtain:
\begin{align}
    \frac{p(y=0|x)}{q(y=0|x)} &= \left[1+p_{dp}(x)\right] p_{rn}(x)\label{isw_n}\\
    \frac{p(y=1|x)}{q(y=1|x)} &= 1+p_{dp}(x)\label{isw_p}
\end{align}
where
\begin{align}
    p_{dp} &= p(y=1)p(h > e)\label{p_dp}\\
    p_{rn} &= \frac{p(y=0)}{p(y=0)+p(y=1)p(h > e)}\label{p_rn}
\end{align}
$p_{dp}(x)$ is the \textit{delayed positive} probability, denote the probability that a sample is a duplicated positive; $p_{rn}(x)$ is the \textit{real negative} probability, denote the probability that an observed negative is ground truth negative and will not convert. 

Finally, considering \meqref{is_1} to \meqref{isw_p}, the importance weighed CVR loss function is:
\begin{equation}
    \begin{aligned}
    &\mathcal{L}_{iw}^{n} = -\sum_{(x_i, y_i)\in \tilde{\mathcal{D}}}^n y_i\left[1+p_{dp}(x_i)\right] log(f_\theta(x_i))\\
    &+(1-y_i)\left[ 1+p_{dp}(x_i) \right]p_{rn}(x_i)log(1-f_\theta(x_i))\label{emp_loss}
    \end{aligned}
\end{equation}
where $\tilde{\mathcal{D}}$ is the training data drawn from elapsed-time sampling distribution $q(x, y)$.

\subsection{Estimation of Importance Weight (IW)}

The challenge of resolving the delayed feedback problem through importance sampling is that we need to estimate the importance weights $w(x,y)$.

In this work, we decompose $w(x,y)$ into two parts: $p_{dp}(x)$ and $p_{rn}(x)$, according to \meqref{isw_n} and \meqref{isw_p}. More precisely, we estimate these two probability with two binary classifiers. Namely, we train a classifier $f_{dp}$ to predict the probability of being a \textit{delayed positive} (\meqref{p_dp}), and train a classifier $f_{rn}$ to predict the probability of being a \textit{real negative} (\meqref{p_rn}). The model architecture of $f_{dp}(x)$ and $f_{rn}(x)$ is the same as the CVR prediction model. To construct the training dataset, for each sample $(x_i, y_i)$, an elapsed time $e$ is drawn from $p(e|x_i)$. Then, for the $f_{dp}$ model, the delayed positives are labeled as $1$, the others are labeled as $0$; For the $f_{rn}$ model, the observed positives are excluded, then the negatives are labeled as $1$ and the delayed positives are labeled as $0$. In practice, all these needed labels are available in a delayed data stream (for example, delayed by 30 days to ensure label correctness), and the data selection can be achieved by a mask on the loss function, thus we train the $f_{rn}$ and $f_{dp}$ models jointly with a shared network in streaming training.

Importance sampling methods usually suffer from high variance due to the division of two probabilities. Our method is less likely to introduce a large variance. The key is on how the importance weight $\frac{p(y|x)}{q(y|x)}$ is calculated. The high variance of importance sampling is mainly introduced by the large value of $\frac{p(y|x)}{q(y|x)}$ when $q(y|x) \ll p(y|x)$ at some $(x, y)$. However, we estimate the importance weight using the delayed positive rate $p_{dp}$ and the real negative rate $p_{rn}$ (in \meqref{emp_loss}), and these two values are probabilities and bounded within $[0, 1]$.

\subsection{Bias Analysis of Estimated IW}\label{bias}
The importance weighted loss function \meqref{emp_loss} is unbiased using ideal $p_{dp}$ and $p_{rn}$. However, a bias may be introduced due to the estimated importance weights $f_{dp}$ and $f_{rn}$. Through optimizing the loss function \meqref{emp_loss}, and using the estimated $f_{dp},f_{rn}$ instead of ideal $p_{dp},p_{rn}$, the predicted probability $f(x)$ converges to: 
\begin{align}
    f(x) &= \frac{p(y=1|x)}{p(y=1|x)+p_{neg}(x)f_{rn}(x)}\label{gdfm_bias}\\
    p_{neg}(x) &= p(y=0|x)+p(y=1|x)p(h>e|x)\label{pneg}
\end{align}

\begin{proofsketch}
Take partial derivative of \meqref{emp_loss} with respect to $f$, and set the derivative to zero. A detailed proof is given in the supplementary material\ref{supp}.
\end{proofsketch}

From \meqref{gdfm_bias} and \meqref{pneg}, we can draw the following observations, which can guide us to design appropriate elapsed-time sampling distribution $p(e|x)$:

\begin{itemize}
  \item First, if $f_{rn}$ is perfectly correct, we have $f_{rn}=p_{rn}$, then $f(x)=p(y=1)$, thus leading to no bias. However, in practice, $f_{rn}$ is learned through historical data, bias always exists.
  \item Second, the bias is also related to $p(y=1|x)$ according to \meqref{gdfm_bias} and \meqref{pneg}. Therefore, if the absolute value of conversion rate is large, the bias introduced by $f_{rn}$ may be larger.
  \item Last, the sampling distribution $p(e|x)$ can be used to control the bias. If $e$ is long, $p(h>e)$ will be smaller. Thus $p(y=0) + p(y=1)p(h>e)$ will be close to $p(y=0|x)$. $f_{rn}$ will be more close to $1$ since there are few fake negatives.Thus $p_{neg}(x)f_{rn}(x)$ is more close to $p(y=0|x)$. 

\end{itemize}

Therefore, we can control the waiting time(elapsed time) distribution $p(e|x)$ to reduce bias, which is the core to realize the aforementioned trade-off and is the missing part of existing methods.

\section{Experiments\footnote{The code for reproducing our results on public dataset is available at \url{https://github.com/ThyrixYang/es_dfm}}}

To evaluate the proposed model, we conduct a set of experiments to answer the following research questions:

\noindent\textbf{RQ1} How does \modelnameabb{} perform, compared to the state-of-the-art models for the streaming CVR prediction task?

\noindent\textbf{RQ2} How do different choices of elapsed time affect the performance? What is the best elapsed time of the dataset?

\noindent\textbf{RQ3} How do mislabeled samples affect importance weighting methods in streaming training?

\noindent\textbf{RQ4} How does ES-DFM perform in online recommender systems?

\subsection{Datasets}

\begin{table*}[htbp]
    \resizebox{\textwidth}{!}{
    \centering
    \begin{tabular}{ccccccccc}
    \toprule[0.75pt]
      Dataset & \# Users & \# Items & \# categorical feature & \# continuous feature & \# conversions & \# Samples &\# average CVR &\# log period \\
    \hline
    Criteo Dataset & - &5443 & 9 & 8 & 3619801 & 15898883 & 0.2269 &60 days\\
    \mdata& 0.25 billion & 0.8 billion & 12 & 10& 0.32 billion& 9.8 billion & 0.03273 & 14 days\\
    \bottomrule[0.75pt]
    \end{tabular}}
    \caption{Statistics of Criteo and \mdata.}
    \label{Taobao_stat}
\end{table*}

\begin{table*}[t]
\resizebox{\textwidth}{!}{
\begin{tabular}{lllllllllllll}
\toprule
\multirow{2}{*}{Method} & \multicolumn{6}{c}{Criteo Dataset} & \multicolumn{6}{c}{\mdata} \\ \cmidrule(l){2-13} 
        &   AUC& PR-AUC  &NLL   & \textit{R-AUC}  & \textit{R-PR-AUC}  & \textit{R-NLL} & AUC& PR-AUC  &NLL   & \textit{R-AUC}  & \textit{R-PR-AUC}  & \textit{R-NLL} \\ \midrule
         Pre-trained & 0.8307  & 0.6251  & \underline{0.4009}  & -0.9212 & -0.2058  & \underline{0.2139} & 0.8731  & 0.6525  &  0.1156  &  -1.0374  & -0.5217   & -0.2419  \\ \midrule
         Vanilla  &  \underline{0.8376}  & 0.6288  & 0.4047  & \underline{0.0000}  & 0.0000  & 0.0000 &  0.8842  & 0.6645  &  0.1141  & 0.0000  &  0.0000 & 0.0000 \\ \midrule
         Oracle*   & 0.8450  & 0.6469  & 0.3868  & 1.0000  & 1.0000  &1.0000  & 0.8949  &  0.6875 &  0.1079  &  1.0000  &1.0000    &1.0000  \\ \midrule
        DFM  & 0.8132  & 0.5784  & 1.2599  &  -3.2581 & -2.7833  & -47.645 & 0.8702   &  0.6471  &0.1271    & -1.3084   &  -0.7565  & -2.0968 \\ \midrule
        FSIW   & 0.8290  & 0.6189  & 0.4099  & -1.1432  & -0.5479 & -0.2891 &  0.8735  & 0.6591  & 0.1149 &-0.9971   & -0.2348   & -0.1290  \\ \midrule
        FNC  & 0.8373  & 0.6267  & 0.4382  & -0.0393  & -0.1147  & -1.8646 & \underline{0.8851}  &0.6669   & 0.1142  &  \underline{0.0841} & 0.1043  & -0.0161 \\ \midrule
        FNW   & 0.8373  &  \underline{0.6313} & 0.4033  & -0.0308  & \underline{0.1400}   & 0.0773 &0.8845   &  \underline{0.6672} &  \underline{0.1137} & 0.0280  & \underline{0.1174}  &\underline{0.0645}   \\ \midrule
        \textbf{\modelnameabb{}}   & \textbf{0.8402*}  & \textbf{0.6393*}  & \textbf{0.3924*} & \textbf{0.3560}  & \textbf{0.5799}  & \textbf{0.6831} & \textbf{0.8895*}  & \textbf{0.6762*}  & \textbf{0.1112*}  & \textbf{0.4953} &  \textbf{0.5087} & \textbf{0.4677} \\ \bottomrule
\end{tabular}
}
\caption{\label{tab:main_table}Performance comparisons of proposed model with baseline models on AUC, PR-AUC and NLL metrics. The bold value marks the best one in one column, while the underlined value corresponds to the best one among all baselines. Here, * indicates statistical significance improvement compared to the best baseline measured by t-test at $p$-value of 0.05. \textit{R-AUC}, \textit{R-PR-AUC} and \textit{R-NLL} are relative metrics indicating the improvements within the delayed feedback gap.}
\end{table*}

\begin{figure*}[t]
\centering
    \includegraphics[width=\textwidth]{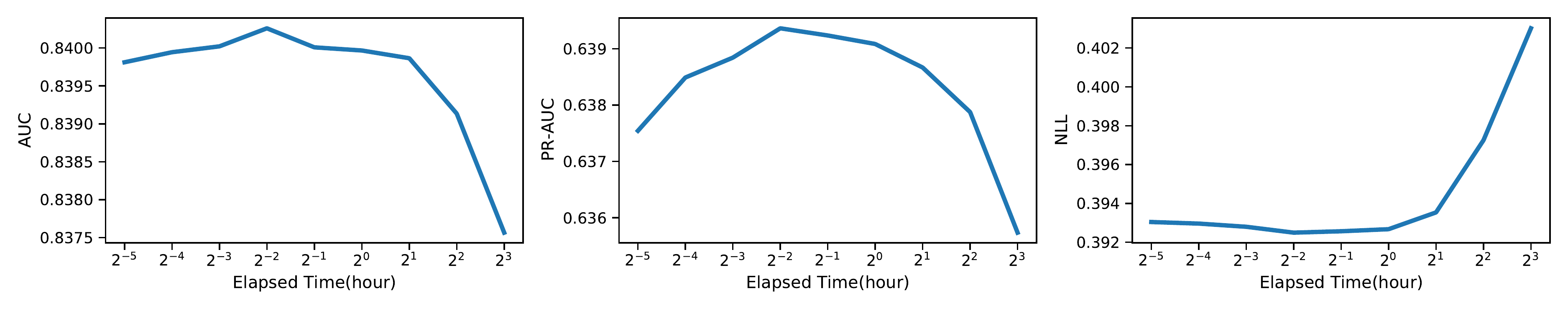}
    \caption{Experiments on the effect of elapsed time on performance. We control the elapsed time by a parameter $c$, which is the value on the $x$ axis.}
    \label{fig:trade-off}
\end{figure*}

\begin{figure*}
     \centering
     \begin{subfigure}[b]{0.33\textwidth}
         \centering
         \includegraphics[width=\textwidth]{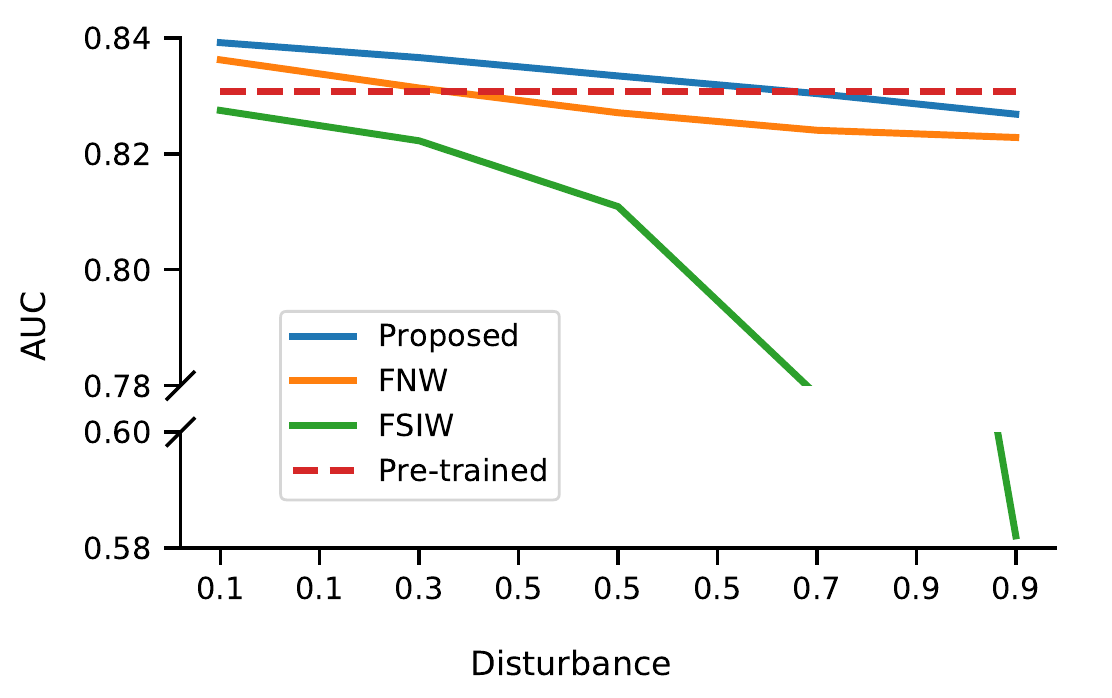}
     \end{subfigure}
     \hfill
     \begin{subfigure}[b]{0.33\textwidth}
         \centering
         \includegraphics[width=\textwidth]{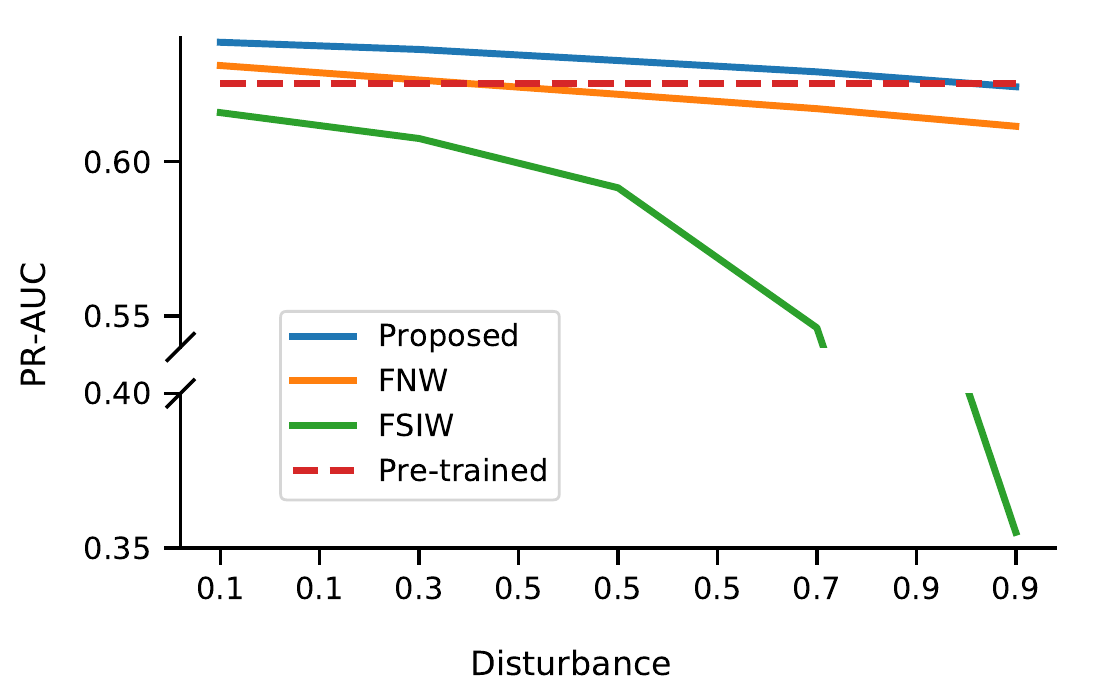}
     \end{subfigure}
     \hfill
     \begin{subfigure}[b]{0.33\textwidth}
         \centering
         \includegraphics[width=\textwidth]{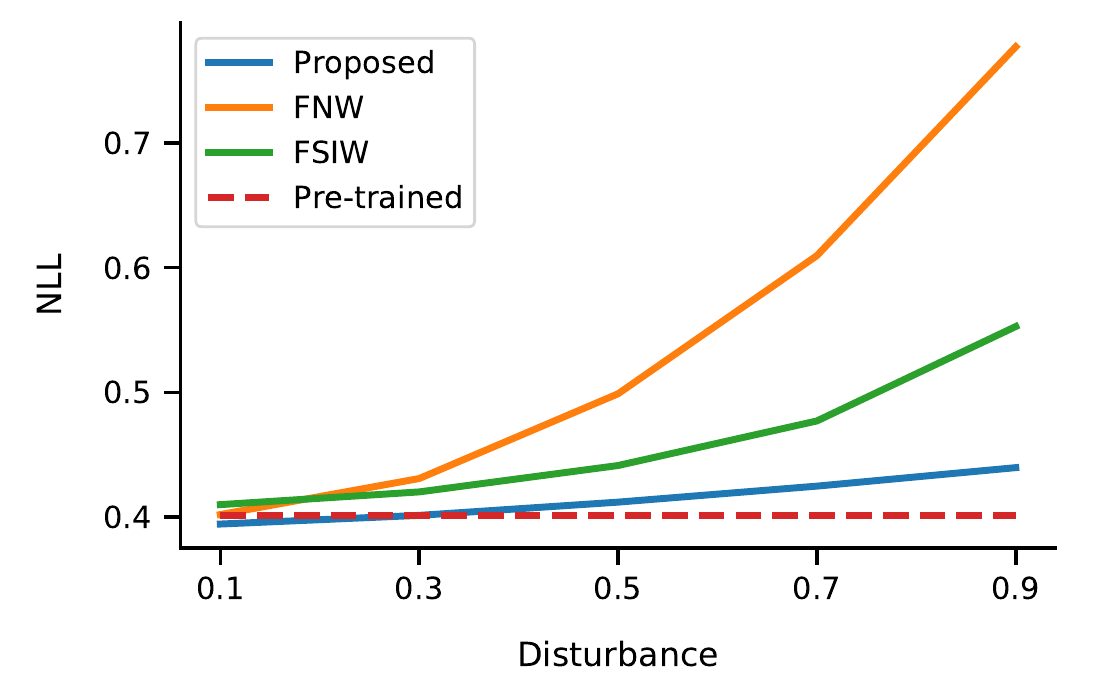}
     \end{subfigure}
        \caption{The experiment on resistance to disturbation. $x$ axis is the disturbation strength which controls the portion of positive samples to be flipped.}\label{fig:disturb}
\end{figure*}

\subsubsection{Public Dataset}
We use the Criteo dataset\footnote{The original url provided by Criteo at 
https://labs.criteo.com/2013/12/conversion-logs-dataset/ is broken, thus we provide a copy at https://github.com/ThyrixYang/es\_dfm . The users should conform to the terms of use from Criteo to use this dataset.} used in~\citet{DFM} to evaluate the proposed method. This dataset is formed by Criteo live traffic data in a period of 60 days, which corresponds to conversions after a click has
occurred. Each sample is described by a set of hashed categorical features and a few continuous features. It also includes the timestamps of the clicks and those of the conversions, if any. The statistics of Criteo dataset is shown in Table~\ref{Taobao_stat}.

\subsubsection{\mdata{}}
We collect $98\times 10^8$ samples in a period of 14 days from the daily click and conversion logs in \mus{} system, which consist of the user and item features with the labels (i.e., click or conversion) for the CVR task. The feature set of an item contains several categorical features and continuous features. The statistics of \mdata{} is shown in Table~\ref{Taobao_stat}.

\subsubsection{Dataset Preprocessing}
We divide both public and anonymous dataset into two parts evenly. We use the first part for model pre-training and achieve a well initialized CVR prediction model. We use the second part for streaming data simulation to evaluate compared methods.

\subsection{Evaluation Metrics}
We adopt three widely used metrics for the CVR prediction task~\cite{ni2018perceive,zhou2019deep,FNW,FSIW}, which show a model's performance from different perspectives. The first metric is area under ROC curve (\textbf{AUC}) which assesses the pairwise ranking performance of the classification results between the conversion and non-conversion samples. The second metric is area under the precision-recall curve (\textbf{PR-AUC}) which is more sensitive than AUC in skewed data like CVR prediction task~\cite{FSIW}. The last metric is negative log likelihood (\textbf{NLL}), which is sensitive to the absolute value of the CVR prediction~\cite{DFM}. In a CPA model, the predicted probabilities are important since they are directly used to compute the value of an impression.

\subsection{Streaming Experimental Protocol}
We have designed an experimental evaluation method for the streaming CVR prediction, which can fully verify the performance of different methods in the online learning settings. In this work, we divide the streaming dataset into multiple datasets according to the click timestamp, each of which contains one hour data. Following the online training manner of industrial systems, the model is trained on the $t$-th hour data and tested on the $t+1$-th hour data, then trained on the $t+1$-th hour data and tested on the $t+2$-th hour data, and so on and so forth. Note that, the training data is reconstructed with fake negatives, while evaluation data is the original data. Therefore, we report the weighted metrics of the evaluation dataset of different hours to verify the overall performance of different methods on streaming data.

\subsection{Compared Methods}

We compare our method with the state-of-the-art methods:

\begin{itemize}

\item[--] \textbf{Pre-trained}: A CVR model without any finetuning.

\item[--] \textbf{Vanilla Finetune Model}: A model finetuned on top of the pre-trained model using the streaming data, which is the baseline method.

\item[--] \textbf{Delayed Feedback Model (DFM)}\cite{DFM}: A model finetuned on top of the pre-trained model using delayed feedback loss.

\item[--] \textbf{Fake Negative Weighted (FNW)}~\cite{FNW}: A model finetuned on top of the pre-trained model using the fake negative weighted loss.

\item[--] \textbf{Fake Negative calibration(FNC)}~\cite{FNW}: A model finetuned on top of the pre-trained model using the fake negative calibration loss.

\item[--] \textbf{Feedback Shift Importance Weighting (FSIW)}~\cite{FSIW}: The pre-trained model will be fine-tuned using the FSIW loss and pre-trained auxiliary model.

\item[--] \textbf{\modelname{} (\modelnameabb{})}: Our proposed method which try to keep the model fresh while introducing low bias.

\end{itemize}

We also reported performance of an \textbf{Oracle*} model: A model finetuned using the ground truth label instead of observed label, assuming the conversion label is known at click time. This is the upper bound of possible improvements, where the delayed feedback problem does not exist. The asterisk* denotes that it's not a baseline method.

\subsection{Parameter Settings}
For fair comparison, all hyper-parameters are tuned carefully for all compared methods. The feature engineering of the numerical features and the categorical features is the same as the settings in the work~\cite{DFM}. Since we mainly discuss the delayed feedback issue in this paper, the model architecture is a simple MLP model with the hidden units fixed for all models with $[256,256,128]$. The activation functions are Leaky ReLU and every hidden layer is followed by a BatchNorm layer~\cite{ioffe2015batch} to accelerate learning. Adam~\cite{kingma:adam} is used as the optimizer with the learning rate of $10^{-3}$. L2 regularization strength is $10^{-6}$. We describe the detailed setting of compared methods in the Supplementary Material \footnote{\label{supp}Some experiment details and discussion are provided at \url{https://github.com/ThyrixYang/es_dfm/blob/master/aaai21_sup.pdf}} due to the page limit.

\subsection{Choice of $p(e|x)$}
The sampling elapsed time distribution $p(e|x)$ can be designed based on expert knowledge and the aforementioned bias analysis. For example, users need more time to consider when buying high-priced products, thus a long waiting time is required. However, the public dataset is anonymized, where information like price-level is unavailable. To verify the effectiveness of introducing $p(e|x)$ in the streaming settings, we perform a simplified implementation of $p(e|x)$. More precisely, we set $p(e=c|x)=1$ where $c$ is a constant, which means $p(e|x)$ degenerates to a Dirac distribution. This brings us two following advantages. First, we can strike the balance between obtaining accurate feedback information and keeping model freshness with a single parameter $c$. Second, we conducted experiments with different $c$ in the public dataset, and the experimental results show that choosing the best $c$ can significantly improve performance. The $c$ is also tuned on the private dataset and we report the best result which is achieved using $c=1$.

\subsection{Standard Streaming Experiments: RQ1}

From~\mtabref{tab:main_table}, we can see that our proposed method improves the performance significantly against all the baselines and achieves the state-of-the-art performance. Moreover, some further observations can be made. First, the performance of DFM and FSIW is worse than the vanilla baseline on both the public and \mdata{}. This is because DFM is difficult to converge, thus failing to achieve a good performance in streaming CVR prediction, and FSIW does not allow the data correction once a conversion took place afterwards, which is important for delayed feedback. Second, in most cases, FNC and FNW perform better than the vanilla baseline. Specially, FNW outperforms the baseline in both PR-AUC and NLL, which is consistent with the results reported in~\citet{FNW}. Third, existing methods show little superior performance in terms of AUC, while our method outperform the best baseline by 0.26\% and 0.44\% AUC scores on the Criteo and \mdata{}, respectively. As reported in~\citet{zhou2018deep}, DIN improves AUC scores by 1.13\% and the improvement of online CTR is 10.0\%, which means a small improvement in offline AUC is likely to lead to a significant increase in online CTR. In our practice, for cutting-edge CVR prediction models, even 0.1\% of AUC improvement is substantial and achieves significant online promotion.

We further analyze the maximum benefit that can be achieved by resolving the delayed feedback problem. The maximum benefit is defined as the performance gap between the oracle model and baseline. Therefore, the goal of any method that tackling delayed feedback problem is to narrow this gap. We report three relative metrics within the performance gap, i.e, Relative-AUC(R-AUC), Relative-PR-AUC(R-PR-AUC) and Relative-NLL(R-NLL). 
As shown in~\mtabref{tab:main_table}, our method can narrow the delayed feedback gap significantly comparing to other methods, and the absolute improvement is larger when the delayed feedback gap is larger.

\subsection{Influence of Elapsed Time: RQ2}

To verify the performance of different choices of elapsed time, we have conducted experiments using different values of $c$ on the Criteo dataset. As shown in~\mfigref{fig:trade-off}, the best $c$ on the Criteo dataset is around 15 minutes, where about 35\% conversions can be observed. Moreover, larger or smaller $c$ will reduce the performance. The performance decreases slowly on smaller $c$, which indicates that the bias introduced by the importance weighting model is small. The performance decreases faster on larger $c$, which indicates that the data freshness matters more when $c$ increase, and a $c$ larger than 1 hour will significantly harm the performance.

\subsection{Experiment on Robustness: RQ3}

In delayed feedback setting, the same sample may be labeled as negative or positive. It is closely related to learning with noisy labels\cite{noisylabel}, where some of the labels are randomly flipped. We hypothesis that a method dealing with delayed feedback problem should not only correct incorrect labels, but also reduce the negative effect of the incorrect labels before they can be corrected or the correction fails (for example, if the weighting model deviate a lot, the bias will be large and correction will fail). Thus we conducted a robustness experiment. We randomly select $d$ portion of all the positive samples in streaming dataset, then swap it's label(and click time and pay time) with a random selected negative one. Note that we do not disturb on the pre-training dataset, so the initial CVR model and the pre-trained importance weighting models are not disturbed. We conducted experiments with different disturbance strength $d$, the results are shown in \mfigref{fig:disturb}. We can see that our method is more resistant to disturbance comparing to FNW and FSIW, and the performance gap is larger when disturbance increases (especially on NLL). We give an intuitive analysis about the weak robustness of FNW and FSIW in the Supplementary Material\ref{supp}.

\subsection{Online Evaluation: RQ4}

We conducted an A/B test in our online evaluation framework. We observed a steady performance improvement, AUC increases by 0.3\% within a 7 days window compared with the best baseline, CVR increases by 0.7\%, GMV(Gross Merchandise Volume) increases by 1.8\%, where GMV is computed by the transaction number of items multiplied by the price of each item. The online A/B testing results align with our offline streaming evaluation and show the effectiveness of ES-DFM in industrial systems.

\section{Conclusion}

The trade-off between the label accuracy and model freshness in streaming training setting has never been considered, which is an active decision of the method rather than a passive feature in offline setting. In this paper, we propose elapsed-time distribution to balance the label accuracy and model freshness to address the delayed feedback problem in the streaming CVR prediction. We optimize the expectation of true conversion distribution via importance sampling under the elapsed-time sampling distribution. Moreover, we propose a rigorous streaming training and testing experimental protocol, which aligns with real industrial applications better. Finally, extensive experiments show the superiority of our approach.

\bibliography{camera_ready_12_15.bbl}

\end{document}